# ThaiCoref: Thai Coreference Resolution Dataset

**Pontakorn Trakuekul**
Department of Linguistics
Chulalongkorn University
pontakorn.tk@hotmail.com

**Wei Qi Leong**
AI Singapore
weiqi@aisingapore.org

**Charin Polpanumas**
Amazon
cebril@gmail.com

**Jitkapat Sawatphol**
School of Information Science
and Technology, VISTEC
jitkapat.s_s20@vistec.ac.th

**William Chandra Tjhi**
AI Singapore
wtjhi@aisingapore.org

**Attapol T. Rutherford**[*]
Department of Linguistics
Chulalongkorn University
attapol.t@chula.ac.th

## Abstract

While coreference resolution is a well-established research area in Natural Language Processing (NLP), research focusing on Thai language remains limited due to the lack of large annotated corpora. In this work, we introduce *ThaiCoref*, a dataset for Thai coreference resolution. Our dataset comprises 777,271 tokens, 44,082 mentions and 10,429 entities across four text genres: university essays, newspapers, speeches, and Wikipedia. Our annotation scheme is built upon the OntoNotes benchmark with adjustments to address Thai-specific phenomena. Utilizing ThaiCoref, we train models employing a multilingual encoder and cross-lingual transfer techniques, achieving a best F1 score of 67.88% on the test set. Error analysis reveals challenges posed by Thai's unique linguistic features. To benefit the NLP community, we make the dataset and the model publicly available at http://www.github.com/nlp-chula/thai-coref.

## 1 Introduction

Coreference resolution is the task of identifying all mentions in a given context that refer to the same real-world entity. This task plays a critical role in enhancing the performance of various NLP applications, including named entity recognition (Dai et al., 2019), dialogue summarization (Liu et al., 2021), question answering (Bhattacharjee et al., 2020), and machine translation (Ohtani et al., 2019). However, accurate coreference resolvers require a large annotated dataset, which can be laborious to obtain.

The development of large annotated datasets has significantly propelled coreference resolution research. Pioneering efforts like the MUC (Message Understanding Conference) (Sundheim, 1995; Chinchor, 1998) and the ACE (Automatic Content Extraction) program (Doddington et al., 2004) lay the groundwork for English coreference resolution. The widely used OntoNotes benchmark (Pradhan et al., 2012) further advanced the field by providing richly annotated corpora across diverse genres in three languages. Following this success, numerous datasets have emerged, both to comprehensively cover various aspects of English (Ghaddar and Langlais, 2016; Webster et al., 2018; Chen et al., 2018; Sakaguchi et al., 2021; Shridhar et al., 2023) and to cater to specific languages (Ceberio Berger et al., 2018; Dobrovolskii et al., 2022; Rohan et al., 2023; Kim et al., 2024).

Despite the expansion of coreference resolution datasets for various languages, Thai text data remains scarce. To the best of our knowledge, only three datasets exist: Han-Coref (Phatthiyaphaibun and Limkonchotiwat, 2023), TH2022News (Theptakob et al., 2023) and KM (Theptakob et al., 2023). However, these datasets are limited in size, domain coverage, and quality control. Han-Coref, which combines news and Wikipedia entries, contains just over 100K tokens. Likewise, TH2022News and KM are both under 20K tokens each, focusing on news and banking domains, respectively. This lack of data impedes the development of competitive Thai coreference resolvers.

To address this limitation, we develop ThaiCoref, a large-scale Thai coreference dataset with over 770K tokens, 44K mentions and 10K entities across four domains. We also leverage a strong baseline architecture and cross-lingual transfer techniques to train a coreference resolver on the ThaiCoref. Moreover, we analyze model errors to understand its performance on Thai text.

We list our contributions below:

- We introduce the largest Thai coreference res-

---

[*]Corresponding author

olution dataset encompassing over 770K tokens, 44K mentions and 10K clusters across four distinct domains, representing a sevenfold increase in size and richer coverage compared to the largest existing dataset.

- We evaluate our dataset's effectiveness by integrating both multilingual and monolingual language models into a baseline architecture and explore the performance of models with and without cross-lingual transfer learning approaches, achieving an F1 score of 67.88%.

- We conduct both quantitative and qualitative analyses of errors. Quantitatively, models may achieve high overall F1 scores but show higher error rates in specific categories. Qualitatively, Thai coreference resolution presents unique challenges due to its isolating nature, lacking morphological cues for grammatical roles. The optional use of relative pronouns and the omission of arguments further introduce ambiguities in identifying mentions and resolving coreferences.

## 2 Related Work

The OntoNotes corpus is a cornerstone dataset providing human-labeled information for various linguistic levels, including syntax, propositions, named entities, word sense and coreference. The corpus contains texts from seven different sources (broadcast conversation, broadcast news, magazine, news wire, telephone conversation, pivot corpus and weblogs) in English, Chinese, and Arabic. Presently, OntoNotes continues to be a widely used benchmark to assess the advancements in coreference resolution models for these languages.

Early coreference resolution models rely on Bidirectional LSTM (Hochreiter and Schmidhuber, 1996) to capture the lexical information within texts (Lee et al., 2017, 2018). However, the advent of Transformer-based encoder models like BERT (Devlin et al., 2019) revolutionize the field. These encoders significantly improve performance across various NLP tasks, including coreference resolution (Joshi et al., 2019).

Pretrained Transformer models have become the dominant architecture for coreference resolution, particularly in English. Works like Kantor and Globerson (2019); Wu et al. (2020); Dobrovolskii (2021); Joshi et al. (2020); Kirstain et al. (2021); Liu et al. (2022); Otmazgin et al. (2023); D'Oosterlinck et al. (2023) and Bohnet et al. (2023) all leverage this architecture to achieve impressive results. Notably, an actor-critic-based neural coreference resolution system (Wang et al., 2021) achieved a state-of-the-art F1 score of 87.5% on OntoNotes English test set.

Anaphora resolution and coreference resolution, while closely related, differ in scope. Anaphora resolution focuses on resolving pronouns and their antecedents, whereas coreference resolution groups all mentions referring to the same entity within a document. Research in Thai anaphora resolution has explored various techniques. One approach utilizes centering theory (Grosz et al., 1995) for analyzing zero anaphora in Thai texts (Aroonmanakun et al., 2009). Kongwan et al. (2022) employs a ranking model to resolve anaphora within Thai Elementary Discourse Units (EDUs). Regarding research on Thai coreference resolution, Phatthiyaphaibun and Limkonchotiwat (2023) introduces the Han-coref dataset and models trained on it via fastcoref package (Otmazgin et al., 2022). Furthermore, Theptakob et al. (2023) proposes a framework for cross-document coreference resolution in Thai, providing a potential solution for languages lacking labeled data resources. Their work presents and utilizes the TH2022News and KM datasets, which include annotations for both within- and cross-document coreference. These aforementioned coreference datasets, however, are substantially smaller and less diverse compared to our ThaiCoref, potentially limiting the generalizability of the models trained on them to a wider range of contexts.

## 3 Annotation Scheme

Our guidelines are primarily derived from Coreference Guidelines for English OntoNotes Version 7.0 (BBN Technologies, 2007). However, we make specific adaptations, including the incorporation of ALIAS links and additions in handling geopolitical entities (GPEs) and consecutive mentions.

### 3.1 Mention Annotation

Mention annotation identifies and tags the spans referring to entities within a document, usually non-generic, non-unspecified, and non-abstract noun phrases. In Thai, subjects and objects of sentences can often be omitted when they are contextually inferred. However, for annotation simplicity, we exclude these omitted arguments. Singletons or entities mentioned only once are also not annotated.

### 3.1.1 Noun Phrase

We mark nouns, proper names, and pronouns, including those in quoted speech, as mentions. We only annotate the longest logical span within a noun phrase. In particular, modifiers are generally included as part of the mention, as long as the modified entity can still refer back to its antecedent. For example, in (1), the modifier "ของออสเตรเลีย" (Australia's) must be considered part of the span.

(1) [ศูนย์วิจัยแอลกอฮอล์และยาเสพติดแห่งชาติของออสเตรเลีย]

[Australia's National Drug and Alcohol Research Centre]

### 3.1.2 Verb Phrase

Verb phrases can also be considered mentions. However, in line with the OntoNotes guidelines, only the head verb of the phrase is annotated for consistency and simplicity. This includes both deverbal nouns formed using the prefixes "การ-" and "ความ-" and noun phrases that refer to the same event but are lexically distinct.

(2) หลังจากนั้นผมก็เดินไป[พูดคุย]$_x$กับเพื่อน ๆ ตามปกติ ในระหว่าง[การพูดคุยนั้น]$_x$ ผมคิดว่าผมยังไม่ถอดเสื้อกาวน์ออก

After that, I walked and [chatted]$_x$ with my friends as usual. During [the conversation]$_x$, I realized that I had not taken off my gown yet.

## 3.2 Link Annotation

Link annotation establishes coreference relationships between mentions within a document. We define three types of links: Identity (IDENT), Appositive (APPOS), and Alias (ALIAS).

### 3.2.1 Identity

The identical (IDENT) links connect mentions that refer to the same entity, concept, or event.

(3) [สมเด็จพระเจ้าโปมาเรที่ 1]$_x$ทรงสละราชบัลลังก์เมื่อปี ค.ศ.1791 แต่[พระองค์]$_x$ ก็ยังคงเป็นผู้สำเร็จราชการแห่งตาฮีตี...

[King Pomare I]$_x$ abdicated in 1791, but [he]$_x$ continued to be the regent of Tahiti...

Impersonal and pleonastic pronouns are not annotated. The primary impersonal pronouns in Thai are "มัน" (it), "เขา" (they) and "ท่าน" (they) (Indrambarya, 2015). The generic mentions are considered only if there exists a specific mention of the same entity they co-refer with within the context.

(4) เขาจะรู้ได้อย่างไรว่า[การพนัน]$_x$เป็นสิ่งไม่ดี ในเมื่อ[มัน]$_x$ถูกกฎหมาย

How does he know [gambling]$_x$ is bad when [it]$_x$ is legal?

Proper names must be treated as a single mention, even if they contain smaller entities.

(5) [ธนาคารแห่งประเทศไทย]

[Bank of Thailand]

Two elements linked by copula verbs, such as "เป็น" or "คือ" (to be) and words functioning similarly, such as "หรือ" (or), are not marked coreferent. In such structures, only the subject is linked to any subsequent mentions.

(6) [สมการข้างต้นนี้]$_x$ก็คือสมการความสัมพันธ์พื้นฐานของสมบัติซิดวลของของไหลที่มีองค์ประกอบคงที่ และจาก[สมการนี้]$_x$...

[The equation above]$_x$ is the fundamental relationship equation for the residual properties of fluids with constant composition, and from [this equation]$_x$...

Temporal expressions, including deictic expressions, such as "ตอนนี้" (now), "วันนี้" (today), and "พรุ่งนี้" (tomorrow), are considered mentions. Dates, months, and years within temporal expressions are not broken down into individual mentions. For example, in the case of "วันที่ 5 ตุลาคม ค.ศ. 1910" (October 5, 1910), the entire span is labelled as a single mention, and its components "วันที่ 5", "ตุลาคม", and "ค.ศ. 1910" cannot be individually labelled as mentions.

(7) จน [พ.ศ. 2514]$_x$ เขาเริ่มโด่งดังเป็นขุนพลเอกของ "ครูเฒ่า" สถิติการชกใน[ปีนี้]$_x$ คือ ชก 9 ครั้ง ชนะ 8 แพ้ 1

Until [1971]$_x$, he became famous as the "Kru Tao's" general. The boxing record for [this year]$_x$ was 9 fights, 8 wins, and 1 loss.

### 3.2.2 Appositive

Appositives (APPOS) connect a head mention to an attribute mention, with the head being the most specific noun phrase and the attribute providing additional details. Attributes usually immediately follow the head or are separated by punctuation, such as commas, colons, dashes, or parentheses. A head mention can have multiple attributes.

(8) [น.ส.สุมัทนา สุขสิงห์]$_{head}$ [อายุ 29 ปี]$_{attribute}$ [ครูผู้สอนวิชาพระพุทธศาสนาให้กับนักเรียนระดับชั้นประถมศึกษาปีที่ 4-6]$_{attribute}$

[Ms.Sumatna Suksing]$_{head}$, [age 29]$_{attribute}$, [a Buddhist Studies teacher for students in grades 4 to 6]$_{attribute}$.

In cases where the head of an appositive phrase is later referred to, the single span containing the entire appositive construction should be linked together.

(9) แพทย์ประจำตัว[[ไมเคิล แจ๊คสัน]$_{head}$ [ราชาเพลงป๊อประดับตำนาน]$_{attribute}$]$_x$ มีความผิดฐานฆาตกรรม [แจ๊คสัน]$_x$ โดยไม่เจตนา

The personal doctor of [[Michael Jackson]$_{head}$, [the legendary King of Pop music]$_{attribute}$]$_x$, has been found guilty of involuntary manslaughter of [Jackson]$_x$.

Job titles can be included as attributes within an appositive phrase.

(10) [ประธานาธิบดีโจ ไบเดน]

[President Joe Biden]

### 3.3 ALIAS

The alias (ALIAS) links connect mentions that refer to the same entity using alternative names, such as aliases, nicknames, or acronyms. These links are particularly used in copula structures or scenarios where the relationship between the mentions does not fit the criteria for IDENT or APPOS links. The main benefit of marking ALIAS is that it ensures all mentions of an entity, regardless of name variation, are accurately captured and linked.

(11) [ทีมที่ชนะเลิศ]$_{head}$ คือ [บราซิล]$_{alias}$

[The winning team]$_{head}$ is [Brazil]$_{alias}$

When the main name is mentioned later and the entire alias construction functions as a noun phrase, the entire phrase should be treated as a single span.

(12) [[ออบรีย์ เดรก แกรห์ม]$_{head}$ หรือมีฉายาในวงการเพลงว่า [เดรก]$_{alias}$]$_x$ เป็นศิลปินเพลงและนักแสดงชาวแคนาดา [เขา]$_x$เป็นที่รู้จักจากการแสดงในบท จิมมี บรูกส์...

[[Aubrey Drake Graham]$_{head}$, or known in the music industry as [Drake]$_{alias}$,]$_x$ is a Canadian musician and actor. [He]$_x$ gained recognition by starring as Jimmy Brooks...

### 3.4 Special Issues

Adjacent spans containing the same amounts of money in different currencies or the same years in different calendar systems are marked as appositives.

(13) [10,000 ปอนด์]$_{head}$ [530,000 บาท]$_{attribute}$

[10,000 pound]$_{head}$ [530,000 baht]$_{attribute}$

We do not link mentions of an organization and a subset of its members.

(14) [สหราชอาณาจักรโปรตุเกส บราซิล และแอลการ์ฟ (United Kingdom of Portugal, Brazil and the Algarves)] เป็นราชาธิปไตยแบบพหุทวีป (pluricontinental) ได้สถาปนาขึ้นโดยการยกระดับ[อาณานิคมของ[โปรตุเกส]คือ รัฐบราซิล] ให้เป็นราชอาณาจักร

[The United Kingdom of Portugal, Brazil and the Algarves] was a pluricontinental monarchy formed by the elevation of [the colony of [Portugal] named State of Brazil] to the status of a kingdom.

GPEs are linked to their governments, including mentions that use the capital city as a metonym for the government, as shown in (15). However, GPEs are not linked to mentions of their populations.

In the Thai language, due to the lack of morphological distinction between noun and adjective forms of GPEs, all GPEs modifying nouns are considered proper nouns. However, when GPEs are used in forms that clearly function as adjectives, such as "อเมริกัน" (American) in "ชาวอเมริกัน" (American people), they can be treated as adjectives.

(15) [รัฐบาลของ[ประเทศไทย]$_x$]$_x$

[[Thailand's]$_x$ government]$_x$

Quantifiers and their associated nouns are treated as one single span.

(16) [สาวน้อย 9 คน]

[9 girls]

Only quantifiers that select the entire set of entities, such as "ทั้งหมด" (all) or "ทั้ง" (both) can establish coreference links between two equivalent sets within a span. Those selecting a subset, on the contrary, cannot link between smaller and larger sets due to the difference in scope.

(17) [[นักเรียน]$_x$ทุกคน]$_x$

[All of [the students]$_x$]$_x$

(18) [[นักเรียน]บางคน]

[Some of [the students]]

Likewise, we do not link mentions that contain two or more distinct entities when these entities are explicitly mentioned as separate individuals.

(19) อาหาร[เกาหลี]และอาหาร[ญี่ปุ่น]เป็นอาหารที่คนไทยนิยมบริโภคมากในตอนนี้ โดยอาหาร[ทั้งสองประเทศ]นั้นมีลักษณะที่ต่างกัน

[Korea's] food and [Japan's] food are very popular among Thai people right now. The food of [both countries] has different characteristics.

Consecutive nominal or pronominal mentions referring to the same entity are annotated as a single span.

(20) [พี่พลอยเขา]$_x$อายุมากกว่าพีคแต่พีคก็รู้สึกว่า[เขา]$_x$เป็นเพื่อนพีคคนหนึ่งเหมือนกัน

[Ploy she]$_x$ is older than Peak, but Peak feels that [she]$_x$ is also one of Peak's friends.

## 4 ThaiCoref Dataset

### 4.1 Sources

Our dataset combines text from two main sources: Thai National Corpus (TNC) (Aroonmanakun et al., 2009) (20,000 sentences) and Thai Wikipedia (10,000 sentences). The TNC portion can be divided into three parts: university essays (5,000 sentences), newspapers (10,000 sentences), and speeches (5,000 sentences), providing a collection across four genres. Note that these sentence counts are estimated based on segmentation using the PyThaiNLP library's sent_tokenize function (Phatthiyaphaibun et al., 2023).

### 4.2 Annotation Process

We recruit nine annotators, primarily university students with a linguistics background. Before commencing the annotation process, the annotators must undergo a comprehensive training, involving a review of the annotation guidelines and a pilot annotation of 3,000 sentences. The pilot results are meticulously analyzed to refine the guidelines, addressing any ambiguities or inconsistencies. Throughout the process, a linguist consultant provides ongoing supervision and guidance. Finally, to guarantee high-quality annotations, we assign three annotators to re-check all 30,000 annotated sentences. This re-checking process reveals that the primary annotation errors stem from narrow span boundaries, tagging inconsistency, generic mention tagging, and annotator negligence.

We utilize Datasaur[1] as our labeling tool. For mention tagging, we opt for character selection to capture entire mentions and mitigate potential parsing errors that may arise from token-based methods, as detailed in section 4.5.

### 4.3 Inter-Annotator Agreement

To ensure consistency in data annotation, we perform an inter-annotator agreement score. We perform this process on a small subset comprising 10% of the full dataset. The process is described as follows. First, we split the rechecked data to obtain the subset. It is important to note that this split is temporary and for calculating the agreement score only. This version of the data subset is denoted as QC1. Second, we perform an additional round of rechecking the same subset using two annotators. The resulting data from this process is referred to as QC2. Finally, to describe the agreement between annotators in the QC1 and QC2 data, we compute the Kappa statistic following the study by He (2007). This Kappa statistic describes the degree of agreement of two sets of annotated samples beyond random chance. We record the final Kappa score at 0.941.

### 4.4 Postprocessing

To optimize both training and evaluation, we apply three postprocessing steps to the annotated data. First, we remove irrelevant elements specific to each document type: titles and metadata from university essays, headlines from newspapers, speaker names and details from speeches, titles, and undesirable sections (e.g., references, sources, and external links) from Wikipedia articles. Secondly, since most newspaper and university essay documents are constructed from multiple articles, we manually split them into individual segments. Lastly, we identify and remove samples with no links and duplicate entries that contain exactly the same content. Following this process, the final dataset comprises 1,614 documents.

---
[1] https://datasaur.ai

### 4.5 Tokenization

For document tokenization, which is essential for dataset statistics and preparing input for the model, we leverage PyThaiNLP's dictionary-based newmm tokenizer (Phatthiyaphaibun et al., 2023). We keep whitespace characters intact since they are crucial for accurately delineating text boundaries in Thai. However, automatic tokenization can sometimes create an error where a mention does not take up a whole token span. For example, the phrase "ในประเทศไทย" (in Thailand) is broken down into "ในประเทศ" (in country) and "ไทย" (Thailand) instead of "ใน" (in) and "ประเทศไทย" (Thailand). To ensure cleaner data, we thus exclude all these erroneous mentions during training and evaluation, resulting in a slight decrease in number of mentions from 44,082 to 42,587 mentions (-3.4%) and entities from 10,429 to 10,143 entities (-2.7%).

### 4.6 Train, Development, and Test Splits

We use a stratified split to allocate data into train, development, and test sets, maintaining consistent proportions from each source. The train set receives 80%, the development set 10%, and the test set 10% of the data. We detail the distribution of our dataset in Table 1.

| Genre | Train set | Dev. set | Test set | Total |
|---|---|---|---|---|
| University Essays | 170 | 21 | 21 | 212 |
| Newspapers | 543 | 68 | 68 | 679 |
| Speeches | 63 | 7 | 8 | 78 |
| Wikipedia | 516 | 65 | 64 | 645 |

Table 1: Dataset Splits

Table 2 provides an overview, including the number of examples, tokens, average document length, entities (chains), average mentions per entity for each data split. The dataset contains 777,271 tokens, 44,082 mentions and 10,429 entities overall.

| Category | Train set | Dev. set | Test set | Total |
|---|---|---|---|---|
| #Documents | 1,292 | 161 | 161 | 1,614 |
| #Tokens | 631,110 | 70,706 | 75,455 | 777,271 |
| Average length | 489 | 439 | 469 | 482 |
| #Entities | 8,267 | 1,073 | 1,089 | 10,429 |
| #Mentions | 35,124 | 4,497 | 4,461 | 44,082 |
| Mentions/Entities | 4.3 | 4.2 | 4.1 | 4.2 |

Table 2: Dataset Statistics

To contextualize the scale of the dataset, table 3 compares the overall statistics of our dataset with three portions of the OntoNotes corpus. Notably, while our dataset has a relatively high token count, it exhibits a low number of mentions per tokens. This suggests a lower mention density compared to the OntoNotes dataset. We hypothesize that this primarily stems from potential over-tokenization due to automatic segmentation in Thai, in contrast to OntoNotes' gold-standard segmentation. Over-tokenization occurs when the tokenizer splits words into smaller units due to limitations in its dictionary. For example, the name "โยโกฮามะ" (Yokohama) is split into four individual tokens: "โย", "โก", "ฮา", and "มะ".

## 5 Experiments Setting

Our primary goal is to investigate the quality of the ThaiCoref dataset in a real machine learning setting. We train coreference resolution models on our dataset and evaluate their accuracy. Following the evaluation practices, we focus solely on evaluating models' ability to identify identical coreferences, which means appositives and aliases are not included.

### 5.1 Models

We choose the simple-but-strong start-to-end (S2E) model (Kirstain et al., 2021) as our baseline. The S2E model offers a significant advantage in terms of memory efficiency compared to the widely used baseline model by Joshi et al. (2019) as it does not require span-level representations. Notably, this model achieves a high F1 score of 80.3 on the OntoNotes English test set with Longformer-Large (Beltagy et al., 2020) as its encoder.

The core of the model lies in pairwise scoring, which combines mention scores $f_m$ and antecedent scores $f_a$ using lightweight bilinear functions over contextualized word embeddings of each mention's start and end tokens. The following equation illustrates how the model calculates the pairwise coreference score $f(c, q)$ between a query mention $q$ and a candidate antecedent $c$, where $\varepsilon$ represents the dummy antecedent.

$$f(c, q) = \begin{cases} f_m(c) + f_m(q) + f_a(c, q) & c \neq \varepsilon \\ 0 & c = \varepsilon \end{cases}$$

The mention scores $f_m(c)$ and $f_m(q)$ assess the likelihood of $c$ and $q$ being valid entities, calculated through a biaffine product involving the representations of the start and end tokens.

$$f_m(q) = v_s \cdot m_{q_s}^s + v_e \cdot m_{q_e}^e + m_{q_s}^s \cdot B_m \cdot m_{q_e}^e$$

| Category | ThaiCoref | English OntoNotes | Chinese OntoNotes | Arabic OntoNotes |
|---|---|---|---|---|
| #Documents | 1,614 | 3,493 | 2,280 | 447 |
| #Tokens | 750K | 1.6M | 950K | 300K |
| #Entities | 10,429 | 44,221 | 35,691 | 10,246 |
| #Mentions | 44,082 | 194,480 | 129,383 | 34,138 |
| Mentions/Tokens | 0.06 | 0.12 | 0.14 | 0.11 |
| Mentions/Entities | 4.2 | 4.4 | 3.6 | 3.3 |

Table 3: Dataset Statistics of the ThaiCoref compared to three portions of the OntoNotes datasets

Here, $m_{q_s}^s$ and $m_{q_e}^e$ denote the start and end token representations of mention $q$. The first two terms capture the individual likelihoods of the start and end tokens being the beginning or ending of an entity mention. The third term measures the compatibility of these tokens as boundaries of the same entity. The vectors $v_s$, $v_e$ and the matrix $B_m$ are trainable parameters.

The antecedent score $f_a(c, q)$ estimates the probability of $c$ being the antecedent of $q$. It is calculated using separate start and end token representations and a combination of four bilinear functions:

$$f_a(c,q) = a_{c_s}^s \cdot B_a^{ss} \cdot a_{q_s}^s + a_{c_s}^s \cdot B_a^{se} \cdot a_{q_e}^e$$
$$+ a_{c_e}^e \cdot B_a^{es} \cdot a_{q_s}^s + a_{c_e}^e \cdot B_a^{ee} \cdot a_{q_e}^e$$

$a_{c_s}^s$ and $a_{q_s}^s$ represent the start representations of $c$ and $q$, respectively, while $a_{c_e}^e$ and $a_{q_e}^e$ represent their end token representations. Each component in the sum measures the compatibility of the spans $c$ and $q$ based on interactions between their boundary tokens.

For a more detailed explanation, we refer readers to the original paper (Kirstain et al., 2021).

In contrast to the original work, we replace the Longformer-Large with three base-sized language models with strong Thai language performance:

1. WangchanBERTa (Lowphansirikul et al., 2021): A RoBERTa-based monolingual encoder model trained on a massive 78.5GB corpus of Thai text.

2. PhayaThaiBERT (Sriwirote et al., 2023): An extension of WangchanBERTa with an expanded multilingual vocabulary and further pretraining on a 156.5GB Thai dataset, improving comprehension of loanwords.

3. XLM-V (Liang et al., 2023): A multilingual language model pretrained on a vast dataset encompassing 100 languages, including 71.7GB of Thai data, featuring an improved vocabulary with a well-balanced representation of each language.

Given the established effectiveness of cross-lingual transfer learning (Pires et al., 2019; Hsu et al., 2019; Conneau et al., 2020; Liang et al., 2023), we explore its potential for Thai coreference resolution by transferring knowledge from the English OntoNotes dataset. To this aim, we experiment with the variations of the XLM-V model:

1. XLM-V-ZERO: We utilize a zero-shot cross-lingual transfer learning approach by finetuning the model solely on English data.

2. XLM-V-TA: We employ target-adapting, further finetuning the XLM-V-ZERO model on the ThaiCoref training data.

3. XLM-V-MIXED: We apply mixed training, finetuning the model on a merged dataset comprising English and Thai training data. To balance the dataset and mitigate potential bias towards English, which is the dominant language, we double the size of the ThaiCoref training set and stratified downsample the English data to get an equal amount of English and Thai data[2].

For fair comparison, we use the same hyperparameters as the original work for all models and adjust the input window size to a maximum of 512 tokens for all RoBERTa-based models. After experimenting with different window sizes, we find that a window size of 384 tokens yields the best performance across all models on the development set.

### 5.2 Evaluation Metrics

We employ the CoNLL-2012 official evaluation metric to assess the performance of our models. This metric calculates the score based on three sub-metrics: MUC (Vilain et al., 1995), $B^3$ (Bagga and Baldwin, 1998), and $CEAF_{\phi 4}$ (Luo, 2005).

---
[2] In addition to this balanced setting, we experiment with different ratios of English and Thai data, including 1:1 and 1:2. We find that this balanced setting achieves the best performance on the development set.

## 6 Experimental Results

We conduct all experiments for up to 150 epochs with early stopping on a single 40GB NVIDIA A100, with training times ranging from 11 to 15 hours per experiment. We save checkpoints every 500 steps, and the model with the best performance on the development set is evaluated on the test set. We demonstrate the evaluation results in Table 4.

The results underscore the significant advantage of cross-lingual transfer learning for Thai coreference resolution. Both XLM-V-TA and XLM-V-MIXED achieve high scores, outperforming models trained solely on Thai data. The results also highlight that, for optimal performance, domain-specific fine-tuning remains crucial, as evidenced by the lower performance of XLM-V-ZERO.

PhayaThaiBERT outperforms WangchanBERTa, likely due to its larger pretraining dataset and expanded vocabulary. Interestingly, among models without cross-lingual transfer, monolingual encoders perform worse than the multilingual XLM-V model, suggesting that XLM-V's multilingual pretraining possibly equips it with a broader and more generalizable knowledge base, benefiting Thai coreference resolution even with less Thai training data.

Our results align with research in Arabic coreference resolution (Min, 2021), where GigaBERT (Lan et al., 2020), pretrained with a massive dataset of Arabic and English text, outperforms AraBERT (Antoun et al., 2020), which had less Arabic training data. Similarly, models with bilingual transfer from English achieve better performance in Arabic coreference resolution.

Despite sharing the same underlying architecture, our models achieve lower F1 scores on the ThaiCoref dataset compared to the model trained and evaluated on the English OntoNotes dataset, which exceeds an 80% F1 score. This discrepancy can be attributed to two factors. First, the ThaiCoref dataset, which comprises the target language data, is notably smaller than the OntoNotes dataset. As a known limitation in supervised machine learning, larger datasets often lead to improved model performance. Second, our underlying encoder architecture (RoBERTa-base) is a less complex compared to the baseline model (Longformer-Large).

## 7 Error Analysis

### 7.1 Quantitative Analysis

To analyze model performance differences, we conduct an automated error analysis using the Berkeley Coreference Analyzer (Kummerfeld and Klein, 2013). Since our dataset lacks syntactic annotations necessary for identifying the head of each mention, we utilize a spaCy-Thai dependency parser[3] to generate dependency trees as input for the analyzer.

The tool analyzes errors by transforming model predictions into gold annotations and categorizes the resulting changes into seven error types: Span Error, where the correct head noun is identified but the boundary is incorrect; Conflated Entities, where multiple distinct gold entities are merged into one; Divided Entity, where a single gold entity is split into multiple entities; Extra Mention, where a mention is not present in the gold data; Extra Entity, where an entity does not correspond to any gold entity; Missing Mention, where a gold mention is absent from the model's output; and Missing Entity, where a gold entity is absent from the model's output. Table 5 details the error distribution for each model. Lower error counts indicate better performance.

XLM-V-ZERO stands out with the lowest conflated entities and extra mentions, possibly due to its underexposure to Thai coreference nuances. Without training on Thai coreference patterns, XLM-V-ZERO may be more cautious, avoiding errors like merging entities or creating extra mentions based on misinterpretations. However, this caution leads to the highest number of missing mentions and entities.

Monolingual encoders show fewer span errors compared to the multilingual model. This suggests that Thai monolingual models better identifying mention boundaries, likely due to their deeper understanding of Thai morphology and syntax from larger Thai pretraining corpus.

Overall, consistent with the observations in (Kummerfeld and Klein, 2013), cross-lingual transfer models, or those performing well, achieve high F1 scores despite higher counts of some error types compared to models with lower F1, indicating a complex interplay between different error types and overall performance.

---
[3]https://github.com/KoichiYasuoka/spaCy-Thai

|  | MUC | | | B³ | | | CEAF$_{\phi 4}$ | | | Avg. |
|---|---|---|---|---|---|---|---|---|---|---|
|  | P | R | F1 | P | R | F1 | P | R | F1 | F1 |
| WanchanBERTa | 77.43 | 54.88 | 64.23 | 68.57 | 43.20 | 53.01 | 64.55 | 38.56 | 48.28 | 55.17 |
| PhayaThaiBERT | 74.37 | 64.33 | 68.99 | 65.02 | 54.15 | 59.09 | 63.10 | 50.28 | 55.96 | 61.34 |
| XLM-V | 73.06 | 68.16 | 70.52 | 62.85 | 58.47 | 60.58 | 64.53 | 53.73 | 58.64 | 63.25 |
| XLM-V-ZERO | 71.14 | 39.75 | 51.00 | 65.88 | 29.32 | 40.57 | 53.43 | 33.67 | 41.31 | 44.30 |
| XLM-V-TA | 77.23 | 72.79 | **74.94** | 67.10 | 64.12 | **65.58** | 67.05 | 59.61 | **63.11** | **67.88** |
| XLM-V-MIXED | 78.44 | 68.99 | 73.41 | 69.87 | 59.71 | 64.39 | 68.12 | 56.84 | 61.97 | 66.59 |

Table 4: Performance of the models on ThaiCoref test set

|  | Span Error | Conflated Entities | Divided Entity | Extra Mention | Extra Entity | Missing Mention | Missing Entity |
|---|---|---|---|---|---|---|---|
| WangchanBERTa | **53** | 236 | 347 | 158 | **50** | 409 | 442 |
| PhayaThaiBERT | 61 | 309 | 365 | 208 | 99 | 381 | 288 |
| XLM-V | 84 | 337 | 342 | 264 | 103 | 350 | 219 |
| XLM-V-ZERO | 89 | **175** | 345 | **90** | 88 | 668 | 512 |
| XLM-V-TA | 79 | 247 | **271** | 204 | 132 | **282** | **211** |
| XLM-V-MIXED | 74 | 237 | 283 | 153 | 105 | 351 | 247 |

Table 5: Automatically identified errors in ThaiCoref test set

## 7.2 Qualitative Analysis

To better understand the error patterns in Thai coreference resolution, we also conduct a qualitative error analysis on the prediction result of XLM-V-TA on the development and test set. We focus on errors specific to the Thai language rather than general coreference challenges like pleonastic antecedents, distant antecedents, and limitations of world knowledge.

### 7.2.1 Morphological Ambiguities

Thai, as an isolating language, relies on word order and context rather than inflectional morphemes to convey grammatical roles. For example, to indicate plurality, Thai uses strategies, such as repeating nouns, using numeral classifiers, prepending nouns with collective nouns like "พวก" (group), or relying on context (Iwasaki and Horie, 2005). While this simplifies word structure, it complicates coreference resolution. Without morphological cues, models must consider all potential candidates, regardless of agreement in features like number, case, gender or animacy, increasing complexity and potential errors.

Given example (A), the model erroneously links the plural mention "ดาบส" (hermits) to the distant singular mention "พระบิดา" (the father) instead of the closer plural mention "ดาบสเหล่านี้" (these hermits) due to the absence of a plural morpheme in the query mention.

(A) เมื่อ∅กลับไปถึง∅ก็แสดงวิชาที่ตนเองได้ร่ำเรียนมาให้[พระบิดา]$_p$จน∅พอพระทัยอย่างมาก...แต่[ดาบสเหล่านี้]$_g$เหาะได้ทุกคน ด้วยความสงสัย∅จึงเข้าไปสอบถามถึงเหตุ และขอให้[ดาบส]$_{gp}$สอนวิชานี้ให้กับตน[4]

When ∅ returned, ∅ showed the skills that they had learned to [the father]$_p$, until ∅ was greatly satisfied...but [these hermits]$_g$ could all fly. Out of curiosity, ∅ approached to inquire about the reason and asked [the hermits]$_{gp}$ to teach him this skill.

The lack of embedded morphemes becomes particularly evident with words that have identical surface forms but multiple grammatical functions. For instance, "ใหม่" in example (B) can function as both an adjective meaning 'new' and an adverb meaning 'newly'. The resolver thus mistakenly parses the phrase as meaning 'calling the temple's new name' instead of 'calling the temple's name newly', interpreting "ใหม่" as an adjective and including it in the mention span.

(B) พระครูสิริสาครวัฒน์ซึ่งเป็นลูกศิษย์ จึงได้ทำหน้าที่พัฒนาและเรียกชื่อ[[วัด]$_{gold}$ใหม่]$_{pred}$ว่า "วัดศรีเมือง"

Phrakru Sirisakornwat, who is a disciple, thus takes on the responsibility of developing and calling [the temple's name]$_{gold}$ newly as "Wat Sri Mueang".

Furthermore, the same pronoun in Thai may encode different grammatical cases, animacy, numbers or genders depending on the context (Iwasaki and Horie, 2005). For example, "เรา" can mean 'I', 'we', or 'you'; "ท่าน" may refer to 'he', 'she',

---

[4] The following examples use ∅ for omitted subjects or objects and subscripts: $g$ for gold coreference entities, $p$ for system-predicted coreference entities, $gold$ for gold mentions, and $pred$ for system mentions

'they', or 'you'; and "ตน" can denote 'he', 'she', 'they' or 'it'. In example (C), the model incorrectly identifies the distant plural mention "เบลล์กับทางแบงค์" (Bell and the banks) as the antecedent for the pronoun "เรา" (I) instead of the correct singular antecedent "เบลล์" (Bell).

(C) ตอนนี้[เบลล์กับทางแบงค์]$_p$ก็พยายามไกล่เกลี่ยประนอมหนี้กันอยู่...เพราะว่ามันเหมือนทุกคนรู้ว่า[เบลล์]$_g$เป็นใครทุกคนรู้ว่า[เรา]$_{gp}$มีชื่อเสียง...

Currently, [Bell and the banks]$_p$ are trying to mediate and negotiate the debts... because it seems like everyone knows who [Bell]$_g$ is, everyone knows that [I]$_{gp}$ am famous...

These examples illustrate the distinctive challenges posed by Thai morphology for coreference resolution models. Overcoming these challenges requires techniques that can effectively handle the absence of morphological cues, general pronouns, and words that have different parts of speech but have the same surface form.

### 7.2.2 Syntactic Ambiguities

The flexible syntactic structure of Thai presents challenges for coreference resolution. Thai sentences can be grammatically correct and meaningful even when certain elements are omitted, causing ambiguities in identifying mention boundaries and referents.

Firstly, the use of relative pronouns like "ที่" or "ซึ่ง" (that/which) is optional when introducing relative clauses, which can result in syntactic ambiguity. In (D), the model misinterprets the sentence as 'Mr. Chanwit Thapsuphan, Deputy Secretary General of the PEC, acted as Secretary General of the PEC', treating "ปฏิบัติหน้าที่เลขาธิการ กช." (acting as Secretary General of the PEC) as the predicate rather than a relative clause modifying the subject.

(D) [[นายชาญวิทย์ ทับสุพรรณ รองเลขาธิการคณะกรรมการส่งเสริมการศึกษาเอกชน กช.]$_{pred}$ ปฏิบัติหน้าที่เลขาธิการ กช.]$_{gold}$ เปิดเผยภายหลังการประชุม กช. ว่า...

[[Mr. Chanwit Thapsuphan, Deputy Secretary General of the Private Education Commission (PEC)]$_{pred}$ acting as Secretary General of the PEC]$_{gold}$, revealed after the PEC meeting that...

On the other hand, the model can misinterpret a clausal complement as a relative clause. In (E), it incorrectly interprets "มีอาการดังกล่าว" (to have these symptoms) as a relative clause modifying the noun "คน" (them) instead of a clausal complement of the verb "ทำให้" (to make).

(E) สาเหตุที่ทำให้[[คน]$_{gold}$มีอาการดังกล่าว]$_{pred}$น่าจะมาจากการเลี้ยงดูของครอบครัว

The reason that makes [[people]$_{gold}$ have these symptoms]$_{pred}$ is probably due to the family upbringing.

Thai coreference resolution is further complicated by the frequent omission of arguments. In such cases, the model must infer the missing element from context, which is not always straightforward. In example (F), the subject of the second sentence, referring to "พระบิดา" (the father), is omitted. The model mistakenly identifies "พระบิดา" as the antecedent for "ดาบส" (the hermit) rather than "ธรรมเสนกุมาร" (Prince Thammasen). If the subject is explicit, it may prevent nonsensical predictions like 'the father let the father sit on the lap'.

(F) เมื่อ∅ไปถึงวัง [พระบิดา]$_p$เห็น[ธรรมเสนกุมาร]$_g$ใส่ชุดดาบสก็เกิดความเลื่อมใสเป็นอันมาก ∅ให้[ดาบส]$_{gp}$นั่งบนตัก สอบถามถึงเรื่องของบุตรคนอื่นๆ...

When ∅ arrived at the palace, [the father]$_p$ saw that [Prince Thammasen]$_g$ was wearing a hermit's costume, greatly filled with admiration. ∅ let [the hermit]$_{gp}$ sit on the lap and inquired about the other sons..."

These errors emphasize the complexities Thai syntax poses for coreference resolution. Optional relative pronoun usage and argument omissions can introduce ambiguities. Robust Thai coreference resolution models must incorporate techniques that adeptly handle these intricacies.

## 8 Conclusion

We introduce the largest dataset annotated for Thai coreference resolution, along with a detailed annotation scheme. We leverage our dataset to develop a state-of-the-art Thai coreference resolution model through a cross-lingual approach. Our findings prove the benefits of transferring knowledge from English to Thai for coreference tasks and emphasize the necessity of domain-specific adaptation for optimal performance. Moreover, our error analysis reveals how certain features of the Thai language, such as its isolating nature and flexible syntactic structure, can challenge the model's coreference resolution capabilities.